\documentclass[10pt, a4paper]{article}

\usepackage{lrec-coling2024} % this is the new style

\usepackage{amsmath}
\usepackage{times}
\usepackage{latexsym}
\usepackage{graphicx}
\usepackage{listings}
\usepackage{arydshln}
\usepackage{amsfonts}
\usepackage{multirow}

\title{Leveraging Linguistically Enhanced Embeddings for Open Information Extraction}

\name{Fauzan Farooqui$^*$ \qquad  Thanmay Jayakumar$^*$ \qquad  Pulkit Mathur \qquad  Mansi Radke} 
\address{Department of Computer Science and Engineering\\ Visvesvaraya National Institute of Technology, India \\
       \{ fauzanfarooqui7, thanmayjayakumar, pmathur2001 \}@gmail.com, mansi.radke@cse.vnit.ac.in } 
\abstract{
Open Information Extraction (OIE) is a structured prediction (SP) task in Natural Language Processing (NLP) that aims to extract structured $n$-ary tuples - usually subject-relation-object triples - from free text. The word embeddings in the input text can be enhanced with linguistic features, usually Part-of-Speech (PoS) and Syntactic Dependency Parse (SynDP) labels. However, past enhancement techniques cannot leverage the power of pretrained language models (PLMs), which themselves have been hardly used for OIE. To bridge this gap, we are the first to leverage linguistic features with a Seq2Seq PLM for OIE. We do so by introducing two methods - Weighted Addition and Linearized Concatenation. Our work can give any neural OIE architecture the key performance boost from both PLMs and linguistic features in one go. In our settings, this shows wide improvements of up to 24.9\%, 27.3\% and 14.9\% on Precision, Recall and $F1$ scores respectively over the baseline. Beyond this, we address other important challenges in the field: to reduce compute overheads with the features, we are the first ones to exploit Semantic Dependency Parse (SemDP) tags; to address flaws in current datasets, we create a clean synthetic dataset; finally, we contribute the first known study of OIE behaviour in SP models.
 \\ \newline \Keywords{Information Extraction, Language Models, Structure Prediction} }

\begin{document}

\maketitleabstract
\def\thefootnote{*}\footnotetext{These authors contributed equally}\def\thefootnote{\arabic{footnote}}
\section{Introduction}
\label{introduction}
The Open Information Extraction (OIE) task involves extracting structured knowledge from natural-language text - usually as a set of triples. For example, for the sentence \textit{The cat sat on the mat.} the OIE triple is: \textit{(cat;sat;mat)}. Unlike Closed Information Extraction (CIE) that relies on a given ontology, OIE can generalize to multiple domains.

Traditionally, OIE was performed using rule-based or statistical methods (\citealp{10.5555/1614164.1614177}; \citealp{mausam-etal-2012-open}; \citealp{angeli-etal-2015-leveraging}; \citealp{10.1145/1999676.1999697}; \citealp{saha-mausam-2018-open}; \citealp{gashteovski-etal-2017-minie}). Presently, neural methods have been explored  (\citealp{stanovsky-etal-2018-supervised}; \citealp{kolluru-etal-2020-imojie}; \citealp{cui-etal-2018-neural}). Recently, the first comprehensive survey on neural OIE was presented by \citep{zhou2022survey}. It suggests that though neural models do improve on rule-based approaches, the improvement is not as significant if they rely on training data bootstrapped from rule-based systems. This is a notable contrast to other NLP tasks, such as machine translation, that show large performance gains by neural models. This occurs due to the ambiguity in deciding which OIE outputs are 'correct' - suggesting the need for gold training data for neural OIE. Our work addresses this too.

Current research in OIE focuses on improving neural models in either of two main approaches: generative, or discriminative. Generative (Seq2Seq) models produce the tuples as a sequence, conditioned on the input sentence and optionally on a possible predicate. On the other hand, discriminative (tagging-based) models tag each token in the sentence, usually a BIO scheme for arguments and predicates. However, they are limited to extracting one tuple for a predicate, though there may be more than one for that predicate. Moreover, such models cannot extract implicit facts. A detailed comparison can be found in the survey \cite{zhou2022survey}. 

Explicit structural or syntactic information, like a Part-of-Speech (PoS) tag or Dependency Parse (DP) head, has been shown to be useful, but they have not been investigated clearly \citep{zhan2020span}. We show how we can take advantage of such task-independent structure from within the sentence to improve on OIE, a structured prediction (SP) task.

SP tasks focus on finding useful information from text (making them similar in nature), like relation or event extraction, named entity recognition, semantic role labeling and OIE. Very recent research - TANL \cite{tanl} and DeepStruct \cite{wang-etal-2022-deepstruct} 
 - shows that training a single generative model for such tasks achieves SOTA performance on many of them. Standardising all such tasks to one format and training on all of them together has been shown to boost the model's ability to learn to extract task-dependent structure from the input sentence for a required task. We build upon these findings too.

Lastly, Pretrained Language Models (PLMs) have become immensely popular in the NLP community majorly due to their transfer learning capabilities. Though fine-tuning PLMs has become the new paradigm of NLP, few works have explored this for OIE. In fact, no such attempt has been done for leveraging Seq2Seq PLMs (like T5 \citep{JMLR:v21:20-074} or BART \cite{lewis-etal-2020-bart}) in the generative approach to OIE. We address this in our work.

This paper's major contributions can be summarized as follows:
\begin{itemize}
    
    \item To demonstrate the usefulness of linguistic structure in boosting performance for the OIE task, we propose two distinct novel word embedding enhancement techniques - Weighted Addition and Linearized Concatenation - that increase performance by upto 24.9\%, 27.3\% and 14.9\% on Precision, Recall and F1 scores over the baseline. We are thus the first to successfully integrate features with a PLM (T5) in OIE, while also being the first to exploit Seq2Seq PLMs for the generative approach to OIE. We believe this to be an important direction in the field, as this can give any neural OIE architecture the power of both PLMs and linguistic tags in one go.
    
    \item We empirically study the effects of using three important word-level linguistic information from the sentence alone: PoS, Syntactic DP (SynDP) and Semantic DP (SemDP) tags. We are the first to exploit SemDP tags, which is also the strongest contender among single linguistic features. They reduce computing overheads by using 72\% less tags compared to its SynDP counterpart, while maintaining the same performance boost. We thus believe SemDP to be a crucial novel step for incorporating useful, scalable linguistic features. 
    
    \item We contribute a synthetic dataset (built from ClausIE) that boosted performance by 73.7\% and 37.9\% on Recall and F1 scores over the Seq2Seq version of the current largest annotated dataset (LSOIE), the latter which we show to be largely unclean and flawed. We believe researchers in the field will find this to be an integral resource, which includes extracted linguistic tags and processed LSOIE outputs too.
    
    \item We are the first to study how a model trained on all other SP tasks, TANL, affects OIE performance, contributing novel insights along the wider SP research direction.
    
\end{itemize}

\section{Related Work}
\label{related_work}
In line with our core contributions, this section describes past work that takes advantage of linguistic structure for SP tasks or trains a single model on all such tasks.

Work demonstrating the advantages of multi-task SP is very recent. TANL \citep{tanl} treats various SP tasks as a general translation task, by formatting the output for each task as an augmented target language using a Seq2Seq PLM, T5. TANL is the first such model that generalises on SP tasks, and does it successfully. However, it does not address the OIE task. We bridge this gap by extending TANL's format for OIE, and fine-tune both T5 and TANL models on it. Details of the format and the dataset creation is described in Section \ref{dataset}.

DeepStruct \citep{wang-etal-2022-deepstruct} performs task-agnostic structural pretraining by formatting various structured prediction tasks, including OIE, as triples, treating the tuple format as the structure itself. The authors fine-tune on an auto-regressive Language Model (LM) called GLM \citep{du-etal-2022-glm}, citing its better performance than T5 on text summarization, a task similar to structured prediction. During their structural pretraining, they train on a subset of the large OPIEC dataset \cite{gashteovski2019opiec}. However, they benchmark on evaluation frameworks and datasets (like OIE2016 \cite{stanovsky2016creating}) that have shown to be less accurate and noisy (\citealp{kolluru-etal-2020-imojie}, \citealp{bhardwaj-etal-2019-carb}). We also believe that the OPIEC dataset is not suitable for direct Seq2Seq tasks, as we describe later on in this section. 

%At the time of writing, their code and models have not been made publicly available. % A smaller size LM is fine for one-task, low-resource, fine-tuning, etc.
\begin{figure*}[h]
    \centering
    \includegraphics[scale=0.8]{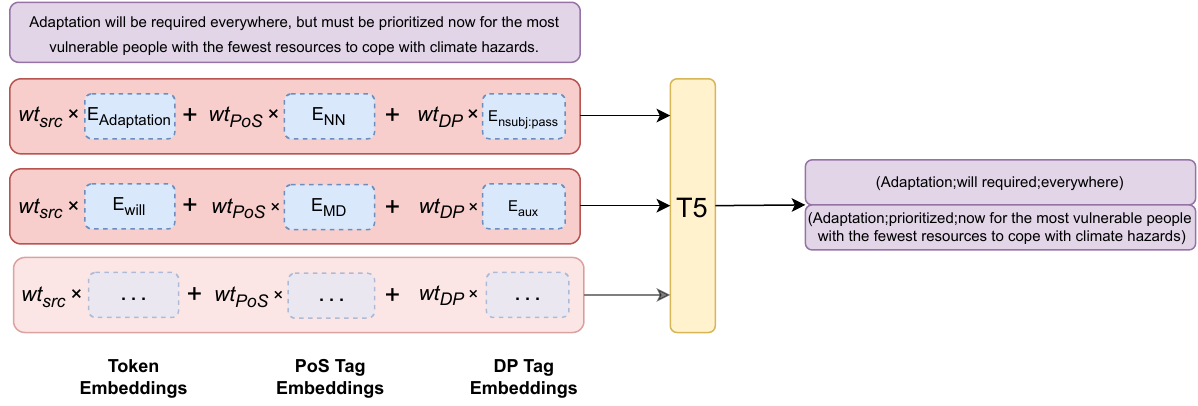}
    \caption{Structure Embedding Addition (Source sentence from the United Nation's website)}
    \label{figadd}
\end{figure*}

While successfully modeling structured prediction as a Seq2Seq task, the above methods rely on information already apparent from the output. Though the PLM may have learned some latent linguistic structure, these models do not explicitly take advantage of such information present solely in the input sentence.

Past work has looked towards including linguistic features such as Part-of-Speech (PoS) tags and Syntactic Dependency Parse (DP) features for the OIE task.

Older works such as Stanford OpenIE \cite{angeli2015leveraging} use handcrafted rules based on DP tags. ClausIE \cite{del2013clausie} uses syntactic information like clause type, PoS and DP roles to extract tuples in a rule-based framework. Its good performance by exploiting solely such linguistic information makes it very promising to use such structure in neural OIE too. We train models on ClausIE's outputs and compare performance to the LSOIE labels. 
MinIE \cite{gashteovski-etal-2017-minie} shifts information from ClausIE's extractions to tuple annotations. OPIEC is the largest OIE dataset built from MinIE's outputs on the Wikipedia database. Though meta-data rich, its special SpaTe format cannot be directly used by existing OIE approaches. Hence, whether DeepStruct still achieves comparable performance on recent OIE benchmarks like CaRB, needs to be verified. 

With the advent of neural networks, RnnOIE \cite{stanovsky-etal-2018-supervised}, SenseOIE \cite{roy-etal-2019-supervising} and SpanOIE \citep{zhan2020span} explore concatenation of such feature embeddings along with source word embeddings.
% RnnOIE: 5:300
SenseOIE conducts a feature ablation study - as a tagging-based model, concatenating other models' tags as part of the input hugely boosts the performance, on which simply  concatenating PoS and syntactic role tags further improves the model. However, embeddings from further related words (parent, left-child, right-child) on the dependency tree have very little effect, noting that simple concatenation is not the best method.
%SenseOIE: 25:300
%SpanOIE: 10/20:100
SpanOIE's results suggest that concatenating the syntactic head's embedding does not help the model. However, they do note that outputs improved in some cases and point out that a better model can be built with a high-quality training corpus and an accurate DP tagger. This could be because they do not make explicit the syntactic role tags as in SenseOIE. Though SenseOIE does not study the effects of including only the syntactic parent embedding, it can be said that SpanOIE's results still echo the conclusion that the head's word embedding may not be as useful as the syntactic role tag itself. The findings from these papers suggest that including information that is directly relevant to the source word could be enough, and a better way to include so should be explored. We present two such methods in the following section.

In the direction closer to our work, \cite{mtumbuka-lukasiewicz-2022-syntactically} concatenate linguistic tag embeddings to the input vectors, and then use GNNs over these inputs to enrich them using other words related through the dependency tree. Their conclusion echo the results of past work on the benefits of including linguistic information. However, this work doesn't investigate the choice of tag embedding size. More importantly, the embeddings have been taken from PLMs but they have not used the full Seq2Seq architecture. We believe to be the first to incorporate linguistic features while using a PLM for the task.

\section{Method}
\label{method}
Structured prediction (SP) tasks already have outputs that closely model their function in the original sentence. For example, in terms of Parts-of-Speech (PoS), in OIE, main predicate words are verbs. In Relation Extraction, both arguments are nouns. Training the model to explicitly exploit such structure within the input will help generate outputs conditioned on an apt linguistic role for that position in the tuple, and also learn syntactic relations between predicates and arguments for better regressive output.

We suggest two novel embedding frameworks to better combine the source word and its linguistic features' word vectors: their "weighted addition" or "linearized-concatenation". These takes advantage of the structure already present in the sentence, without depending on the task at hand, unlike DeepStruct and TANL. These approaches are explained in this section.
Apart from the structured prediction task, this is also independent of whether the task is formulated as generative or tagging-based. 
% This helps improve the nature of tasks that the model can handle without the need of formatting the output in a required way (which recent models depend on).

% along with DeepStruct's general triple output format that is already apparent in the OIE task. 
% We also compare the usefulness of TANL's model over plain T5 for OIE. 

We use Stanza \cite{qi2020stanza} for obtaining the PoS tags and Syntactic DP (SynDP) tags, and SuPar\footnote{\url{https://github.com/yzhangcs/parser}} for the Semantic DP (SemDP) tags. These tags are detailed in Section \ref{Linguistic_Features}.
% They are further described in Appendix \ref{Linguistic_Features}.

\subsection{Weighted Addition} 

In this setting, we obtained the input word embeddings from the pre-trained model, and added it to the embeddings of the PoS and/or DP tags. The tag embeddings are learnt during training, with their dimensions being the same as the word embedding size of the architecture. These enhanced embeddings are then finally passed on to the model, which outputs the triple extractions (Figure \ref{figadd}). 

Formally, let $S$ be the input sentence of length $N$ words, P be the possible set of PoS tags and $D$ be the possible set of dependency tags. For each word $w_i \in S$, the structurally enhanced word embedding $x_i \langle i \rightarrow 1,2,...,N \rangle$ is as follows:
\begin{align*}
 x_i &= wt_{src} \times emb(w_i) \\ &+ wt_{PoS} \times emb(PoS(w_i)  \\ &+
 wt_{DP} \times emb(DP(w_i))
\end{align*}

where $PoS(w_i) \in P$ and $DP(w_i) \in D$ are respectively the the PoS tag and DP tag of $w_i$; $emb(\cdot)$ is the respective $d$-dimensional embedding, that is either learnt during training time, or pre-trained; $wt_{src}$, $wt_{PoS}$ and $wt_{DP}$ are the fractional weights given to the source, PoS and DP tag embeddings respectively.

Note that we take weight values such that, $wt_{src} + wt_{PoS} + wt_{DP} = 1$ and set all the embeddings $emb(\cdot)$ to have the same dimension, in order to make the addition possible.

A weighted addition would help the source embedding to take explicit account of the word's linguistic function across sentences. Instead of storing such information separately, adding it to the embedding itself would easily shift the word in the embedding vector space to better group around words with similar syntactic properties, while also retaining semantic similarity with other words.

\begin{table*}[h]
\centering
\begin{tabular}{|p{0.31\linewidth}|p{0.31\linewidth}|p{0.31\linewidth}|}
\hline
\textbf{LSOIE Sentence} &
  \textbf{LSOIE Labels} &
  \textbf{Generated Labels}
   \\ \hline
Akerson will also relinquish his chairman role, to be replaced by current director Theodore Solso. &
  (Akerson;will relinquish;his chairman role) &
  (Akerson;will relinquish;his chairman role) (current director Theodore Solso;will replaced;Akerson)
   \\ \hline
Road accidents killed 8,600 on the nation's roads last year. &
  (on the nation's roads last year;killed;8,600) &
  (Road accidents;killed;8,600 on the nation's roads last year)
   \\ \hline
He said the world and the Paralympic movement is aware of the situation in the Ukraine, but the IPC needs to stay true to its mission. &
   (the IPC;needs;to stay true to its mission) (the IPC;stay;true to its mission) &
   (the IPC;should stay;true to its mission)
   \\ \hline
\end{tabular}
\caption{Examples where the original LSOIE (test) data is not clean and how our model (trained on the LSOIE-extracted dataset) gives better extractions despite being trained on it}
\label{tab-LSOIEissues}
\end{table*}

\subsection{Linearized Concatenation}

\begin{figure}[h]
    \centering
    \includegraphics[scale=0.55]{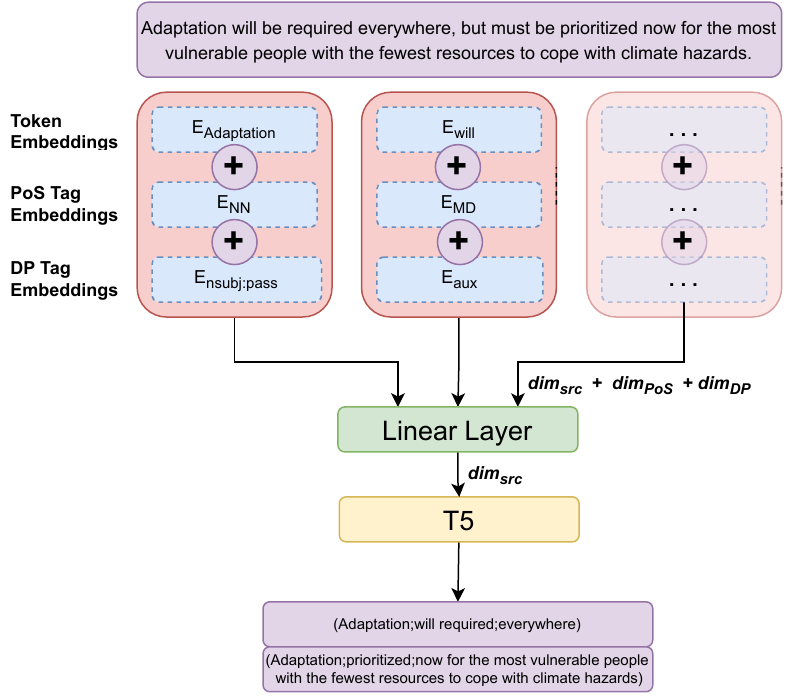}
    \caption{Structure Embedding Concatenation (Source sentence from the United Nation's website)}
    \label{figcat}
\end{figure}

Similar to the previous setting, we obtain the word embeddings from the PLM, and the PoS or DP tag embeddings which are learnt during training. We concatenate both along the embedding dimension. We then introduce a linear layer that converts the elongated embedding dimensions to the source dimension that the model expects. These enhanced embeddings are then passed on to the model to get the tuple predictions (Figure \ref{figcat}).

Formally,
\begin{align*}
 x_i &= g(emb(w_i)\\ 
 &\oplus emb(PoS(w_i)))\\
 &\oplus emb(DP(w_i)))
\end{align*}

where $g(\cdot)$ denotes a linear layer that brings the concatenated size back to the dimensions that the PLM  expects, ie. it is a linear transformation function from $\mathbb{R}^{dim_{src}+dim_{PoS}+dim_{DP}} \rightarrow \mathbb{R}^{dim_{src}}$. Here, $dim_{src}$, $dim_{PoS}$ and $dim_{DP}$ denote the dimensions of the source token, PoS tag and DP tag embeddings respectively.
The remaining notations hold the same meaning as in the previous setting.

Differing from the previous works on simple concatenation, the linear layer would help distribute back the syntactic properties across the embedding vector. The source embedding itself is directly updated in its vector space, unlike the simple shift by WA. This would make the syntactic information be used less explicitly than direct addition, but the back-propagation would make the update more permanent.

\subsection{Learning of tag embeddings}
For learning the PoS and DP tag embeddings, we employ a separate embedding layer that is trained from scratch, as mentioned earlier in this section. The weight update of this layer is done as part of the main transformer (T5) network weight update, and there is no change in the underlying T5 model. Hence, during the training of the entire transformer network, this tag-embedding layer learns to associate the tags with their embeddings such that the overall training loss of the transformer network is minimized.

In the case of Linearized Concatenation, the additional linear layer that is used to redistribute the concatenated source+tag embeddings back to the word embedding dimensions that the PLM expects is used to steer the output with the linguistic information.  

% Although mathematically similar, our empirical experiments show that the methods can give quite different results given specific scenarios, showing that both methods can be considered unique.

\section{Datasets, Processing and Evaluation}
\label{dataset}
LSOIE \citep{solawetz-larson-2021-lsOIE} is the largest human-annotated OIE dataset. Encouraging explicit extractions, it labels each word as an argument/predicate within the BIO scheme, for each predicate. There may be multiple extractions for the same predicate. The order of words in the tuple would then be the same as in the original sentence. They formulate OIE as a tagging-based task and train some models that are benchmarked on their own scorer. It also contains a domain split between the Wiki and Science domains.

LSOIE is large, but it is also quite noisy, which makes it harder to get good scores when benchmarked, especially recall. A subjective analysis of LSOIE (Refer Section \ref{LSOIE_Issues}) shows that our model's output is often better than LSOIE's extractions themselves.  Though LSOIE is the largest human-annotated dataset for OIE, we find that simple cleaning is not enough and the dataset needs to be manually reannotated for triples by trained judges, underscoring the need for higher-quality datasets in OIE. We find an almost instant remedy by training on ClausIE's outputs, cementing our claim that the issue is indeed with LSOIE's dataset.

\subsection{Subjective Analysis}
\label{LSOIE_Issues}

Table \ref{tab-LSOIEissues} shows three sentences from the LSOIE test set. The first example shows that LSOIE's annotations don't teach the model to extract all candidate tuples, which heavily affects recall scores. However, our model has still learned to output both required tuples in this case. The second example shows that LSOIE has wrong extractions where the main subject itself was not extracted, affecting both precision and recall. Again, our model has still learnt to give a complete extraction. The third example shows the redundancy of multiple extractions in the dataset, and reinforces the issue of not extracting all candidate tuples. Such examples are common across the dataset, heavily affecting both precison and recall. In this case, our model has learnt to not be redundant, but has missed extracting information from the first part of the sentence, as such noisy training examples are frequent in LSOIE.

\subsection{Synthetic Dataset using LSOIE}

We convert each example's OIE output from the BIO form to the triple form before training. For creating the sequential triples, we use the tags assigned to them in the dataset: A0-B, A0-I... forms the subject, P-B, P-I... forms the predicate and A1-B, A1-I..., forms the object. Further subjects (A2, A3...) are concatenated with the A1 subject itself. Each extraction is surrounded with the standard ( ) tokens, with the ; token separating the triples. If a sentence has multiple extractions, they are sequentially concatenated to the output. Hence, we finally have an input sequence sentence and an output sequence of triples demarcated with ( ) tokens, which can now be fed to any Seq2Seq architecture. 
We use the LSOIE Wiki split as we believe that this is better suited for OIE, since the examples are not restricted to any domain. Henceforth, we shall refer to this as the LSOIE-extracted dataset.

\subsection{Synthetic Dataset using ClausIE}
ClausIE has largely been a successful clause/rule-based OIE system that extracts facts comprehensively. We find that for the same input sentences, its outputs are almost always better than LSOIE's outputs (that were themselves derived from other sources). As the system is transparent and extracts all expected facts (except implicit facts), we create another dataset from ClausIE outputs to alleviate LSOIE's lack of quality data. 
% The creation of this dataset is described in Appendix \ref{ClausIE_Creation}. Henceforth, we shall refer to this as the ClausIE-extracted dataset.
% Moreover, we also experiment with the TANL format for ClausIE. 

\subsection{Extending TANL's format for OIE}
As mentioned earlier, TANL is trained for other SP tasks but not for OIE. We create a format in line with TANL's aims, Our format also handles multiple extractions per verb-predicate, which TANL can naturally handle. Moreover, we design a strategy to explicitly handle multi-predicate sentences to help the model better output words conditioned on an "expected" predicate before hand. 
% The detailed description of the TANL format can be found in Appendix \ref{Extending_TANL}. 
All the models that were fine-tuned on TANL use this format. Henceforth, we shall refer to this as the TANL-format.

\subsection{CaRB Evaluation}
\label{CaRB_Evaluation}
CaRB \citep{bhardwaj-etal-2019-carb} is an evaluation benchmark for OIE. The dataset was created by crowdsourcing manual annotations for OIE2016's \citep{stanovsky2016creating} sentences. It contributes a comprehensive evaluation framework that fairly takes into account the output styles of diverse OIE systems. CaRB matches relation with relation, and arguments with arguments, which is a much better evaluation criteria than some older benchmarks, such as OIE2016, which serializes the tuples into a sentence and just computes lexical matches. Though CaRB can give AUC and PR curves when confidence scores are provided for each extraction, we do not do so, as we use a generative architecture which doesn't give such direct tag scores.

We note that the LSOIE dataset allows $n$-ary tuples, but OIE benchmarks expect multiple triples to be extracted instead. This would cause models directly trained on LSOIE to score lower on CaRB than other models trained on triple-only datasets. Due to this, we follow the pre-processing stage outlined in Section \ref{LSOIE_Issues}.

\section{Experiments}
\label{experiments}
We create a baseline for each model that trains without any linguistic feature. Thus, our intention isn't to propose a full new architecture to beat SOTA, but rather to give a framework that can help any model, including a SOTA one. Thus, we experiment with three types of datasets (LSOIE-extracted, ClausIE-extracted and TANL-format), each of which forms their own baseline. We demonstrate that our methods help improve performance on each such baseline.

When incorporated, the feature tag embeddings are learned during training. All our models are trained on the LSOIE-wiki split of the dataset, to keep the model independent of domain. We use the pre-trained \texttt{t5-base} version for the T5 model. 

For WA, we assign a fractional weight to the input word embeddings and linguistic feature(s). For concatenation, we fix a embedding size for the linguistic feature(s). We experiment with various such embedding fractional weights and sizes.

As TANL and DeepStruct show state-of-the-art results for structured prediction tasks on Seq2Seq training, we chose the generative approach for OIE. Further, we use TANL's success with the pre-trained T5 model. We train on processed LSOIE datasets and benchmark our models on CaRB.
% Hyperparameters, carbon footprint statement and training times are described Appendix \ref{Experimental_Details}.
\begin{figure*}[h]
    \centering
    \includegraphics[scale=0.35]{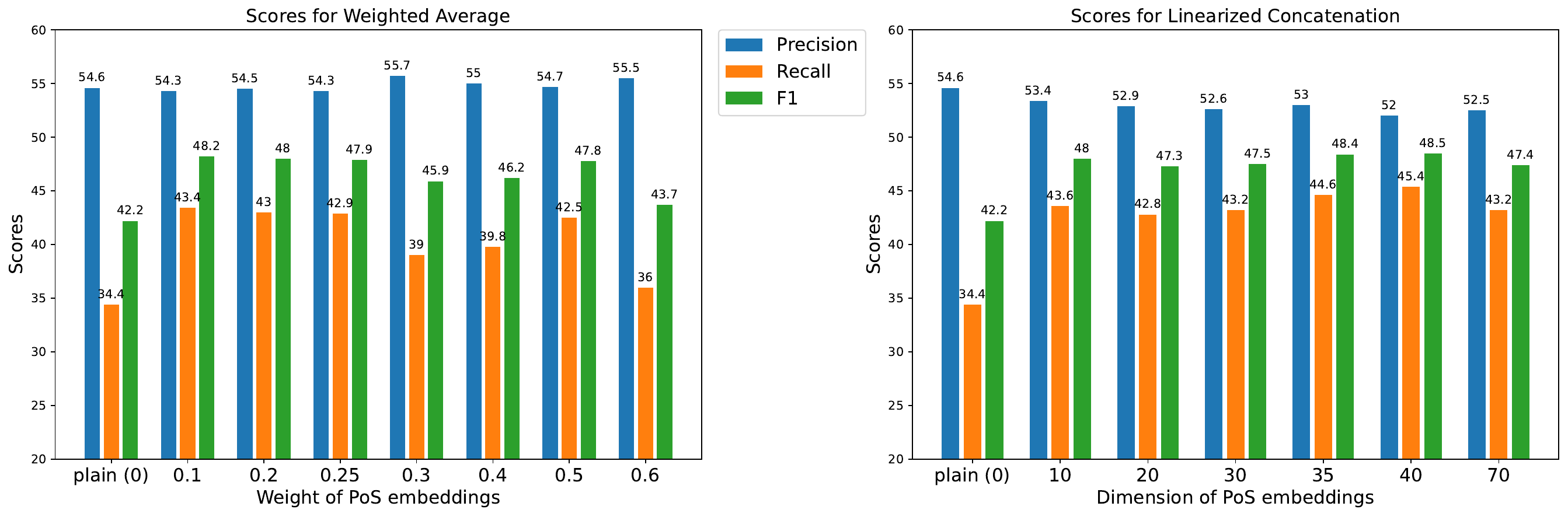}
    \caption{Scores for different $wt_{pos}$ for Weighted Addition (left) and different PoS embedding dimensions for LC (right), trained on the ClausIE-extracted dataset.}
    \label{figLSOIE_2}
\end{figure*}

\begin{figure*}[h]
    \centering
    \includegraphics[scale=0.35]{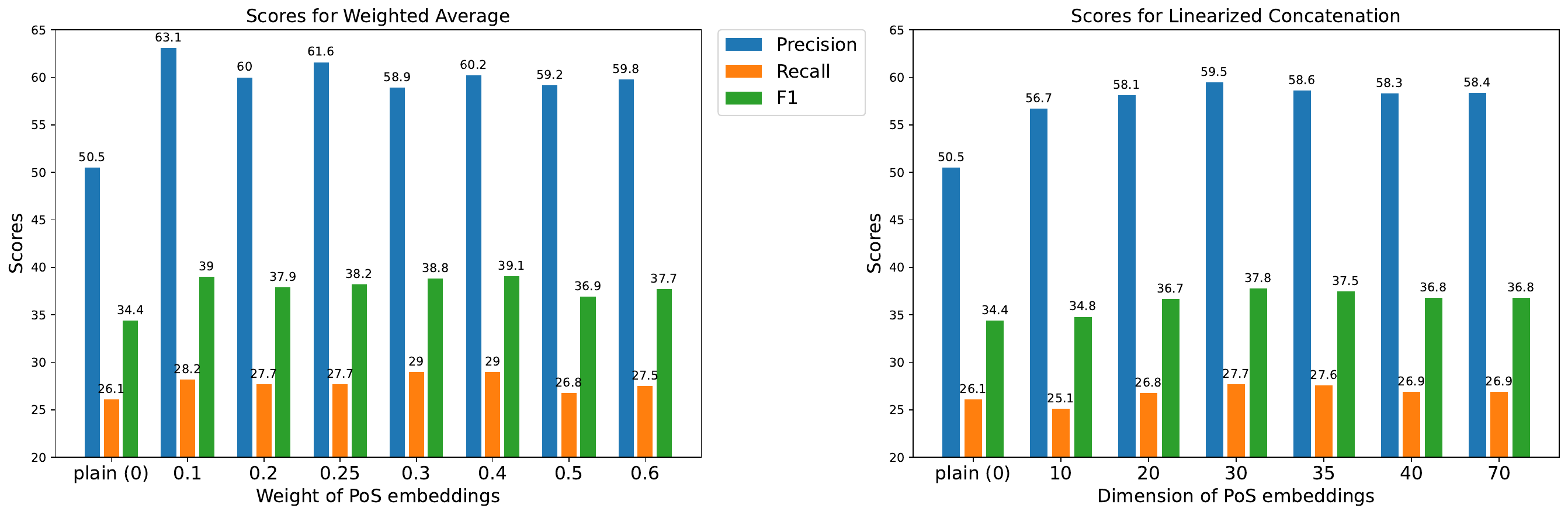}
    \caption{Scores for different $wt_{pos}$ for Weighted Addition (left) and different PoS embedding dimensions for LC (right), trained on the LSOIE-extracted dataset.}
    \label{figLSOIE_1}
\end{figure*}

\subsection{Results}
Table \ref{WA_SCORES_LSOIE} and Table \ref{LC_SCORES_LSOIE} present chosen results for the  Weighted Addition (WA) and Linearized Concatenation (LC) settings respectively on the LSOIE-extracted dataset. Table \ref{WA_SCORES_CLAUSIE} and Table \ref{LC_SCORES_CLAUSIE} present chosen results for the WA and LC settings respectively on the ClausIE-extracted dataset. The dataset baseline being compared to for that model forms the first line of each table (plain implies without linguistic features). 
% Apart from the various main experiments involving linguistic features, datasets and weight/tag-sizes, we conduct other ablation studies which we describe in Appendix \ref{Appendix_Ablations}. 

A wide range of fractional weights and concatenation sizes were tested for the LSOIE-extracted and ClausIE-extracted datasets using the PoS feature, which are presented in Figure \ref{figLSOIE_2} and \ref{figLSOIE_1} respectively. The best results from these models were used to inform our choices of our further experiments. For example, from our figures we see that the best scores are almost always obtained with 0.4 and 0.1 for the fractional weight and 10 and 30 for the concatenation size of the linguistic information respectively. These values were used to inform our choices of further experiments using SemDP and SynDP and we report these values in the tables. We could have done a similar exhaustive analysis for all possible combinations in SemDP and SynDP but that is extremely inefficient (and also has environmental concerns of unnecessary carbon emissions each time) compared to an evidence-based decision of choosing the values. Also, since we mention that training on the ClausIE-extracted dataset (Table 1,3) gave superior scores than LSOIE (Table 2,4) in \textit{all} cases, we limited our experiments for LSOIE.

\subsection{Analysis}

Both our proposed embedding enhancements, WA and LC, provide significant improvements over the plain-embedding model. The best improvement in Precision \textit{P} (+12.6/+24.9\%) happens with adding just 10\% of the PoS tag embeddings (Table \ref{WA_SCORES_LSOIE}) for the LSOIE-extracted dataset. The best improvement in Recall \textit{R} (+9.4/+27.3\%) happens with just concatenating a 30-dimension-size tag of SemDP embedding for the ClausIE-extracted dataset (Table \ref{LC_SCORES_CLAUSIE}). Overall, the best $F_1$ score improvement (+6.3/+14.9\%) happens due to just a 30-dimension-size tag of PoS embedding for the ClausIE-extracted dataset (Table \ref{WA_SCORES_CLAUSIE}).
This indeed points to the importance of enhancing word embeddings with their linguistic information.

\begin{table}[h]
\centering
\resizebox{\columnwidth}{!}{%
\begin{tabular}{cccc|ccc}
$\boldsymbol{wt_{src}}$ & $\boldsymbol{wt_{PoS}}$ & $\boldsymbol{wt_{SynDP}}$ & $\boldsymbol{wt_{SemDP}}$ & \textbf{P} & \textbf{R} & \textbf{F1} \\ \hline
1       & -       & -        & -        & 54.6 & 34.4 & 42.2 \\ \hdashline
0.6     & 0.4     & -        & -        & 55.0 & 39.8 & 46.2 \\
0.6     & -       & -        & 0.4      & 54.4 & 40.1 & 46.2 \\
0.6     & 0.15    & 0.25     & -        & 54.7 & 38.3 & 45.0 \\  
0.9     & 0.1     & -        & -        & 54.3 & 43.4 & 48.2 \\
0.9     & -       & -        & 0.1      & 54.3 & 41.6 & 47.1 \\
\end{tabular}
} 
\caption{P, R and F1 scores for various WA settings. Trained on the ClausIE-extracted dataset.}
\label{WA_SCORES_CLAUSIE}
\end{table}

\begin{table}[h]
\centering
\resizebox{\columnwidth}{!}{%
\begin{tabular}{cccc|ccc}
$\boldsymbol{wt_{src}}$ & $\boldsymbol{wt_{PoS}}$ & $\boldsymbol{wt_{SynDP}}$ & $\boldsymbol{wt_{SemDP}}$ & \textbf{P} & \textbf{R} & \textbf{F1} \\ \hline
1       & -       & -        & -        & 50.5 & 26.1 & 34.4 \\ \hdashline
0.6     & 0.4     & -        & -        & 60.2 & 29 & 39.1 \\
0.6     & -       & 0.4      &          & 60.5 & 29.2 & 39.4 \\
0.6     & 0.15    & 0.25     & -        & 61.1 & 28.9 &	39.3 \\  
0.9     & 0.1     & -        & -        & 63.1 & 28.2 & 39.0 \\
\end{tabular}
} 
\caption{P, R and F1 scores for various WA settings. Trained on the LSOIE-extracted dataset.}
\label{WA_SCORES_LSOIE}
\end{table}

\begin{table}[h]
\centering
\resizebox{\columnwidth}{!}{%
\begin{tabular}{cccc|ccc}
$\boldsymbol{dim_{src}}$ & $\boldsymbol{dim_{PoS}}$ & $\boldsymbol{dim_{SynDP}}$ & $\boldsymbol{dim_{SemDP}}$ & \textbf{P} & \textbf{R} & \textbf{F1} \\ \hline
768     & -       & -        & -        & 54.6 & 34.4 & 42.2 \\ \hdashline
768     & 30      & -        & -        & 52.6 & 43.2 & 47.5 \\
768     & -       & -        & 30       & 53.0 & 43.8 & 47.9 \\
768     & 30      & 30       & -        & 53.5 & 43.7 & 48.1 \\  
768     & 10      & -        & -        & 53.4 & 43.6 & 48.0 \\
768     & -       & -        & 10       & 53.3 & 42.1 & 47.1 \\
768     & 40      & -        & -        & 52.0 & 45.4 & 48.5 \\
\end{tabular}
} 
\caption{P, R and F1 scores for various LC settings. Trained on the ClausIE-extracted dataset.}
\label{LC_SCORES_CLAUSIE}
\end{table}

\begin{table}[h]
\centering
\resizebox{\columnwidth}{!}{%
\begin{tabular}{cccc|ccc}
$\boldsymbol{dim_{src}}$ & $\boldsymbol{dim_{PoS}}$ & $\boldsymbol{dim_{SynDP}}$ & $\boldsymbol{dim_{SemDP}}$ & \textbf{P} & \textbf{R} & \textbf{F1} \\ \hline
768     & -       & -        & -        & 50.5 & 26.1 & 34.4 \\ \hdashline
768     & 30      & -        & -        & 59.5 & 27.7 & 37.8 \\
768     & -       & 30       & -        & 58.2 & 26.5 & 36.4 \\ 
768     & 10      & -        & -        & 56.7 & 25.1 & 34.8 \\
\end{tabular}
} 
\caption{P, R and F1 scores for various LC. Trained on the LSOIE-extracted dataset.}
\label{LC_SCORES_LSOIE}
\end{table}

\textbf{Choice of dataset:}
We observe a general trend that if a jump in one parameter is high, the other parameter's increase isn't as significant. We elaborate on this further. Recall in the original dataset is very limited. Both our synthetic datasets (ClausIE-extracted and TANL-format) give an immediate solution with the large jump in recall scores. Thus, we attribute the low recall scores to the unclean LSOIE extractions (as pointed out earlier). Moroever, both the plain $P$ and $R$ scores of the ClausIE dataset are quite higher than that of LSOIE. However, $P$ sees little to no improvement for the ClausIE dataset.  

This interesting contrast may happen due to the way the model learnt to extract the tuples, as elaborated below. Because LSOIE is unclean, our hypothesis of using linguistic features hugely benefits $P$ in the first place. However, recall observes only a slight help, which is because after the first or second tuple, multi-tuple extractions in the dataset are very noisy and the linguistic features would be confused with them.

On the other hand, the ClausIE extractions are already due to transparent, syntax-based rules. Hence, our linguistic features cannot provide more improvement on it. On the other hand, they help jump up recall because the model has clean multi-tuple extractions to learn to associate the linguistic features with. This shows that the features help span across tuples. This key observation also forms an exciting argument as to why neural OIE systems can do better than clean rule-based systems (when helped by them in the first place). Even though training on a ClausIE dataset may not cover implicit facts or such aspects, it seems that feeding the linguistic structure could have helped in doing so.

Overall, because of either $P$ or $R$ improvement limitations in a dataset, the F1 score doesn't seem so large. However, that atleast one parameter does indeed improve hugely, given the contrasting natures of the datasets, establishes the usefulness of both our methods on any type of dataset.

\textbf{How does the TANL-format help the LSOIE dataset?} The TANL-format on LSOIE show bitter-sweet results: on just the plain model, $R$ comes up by +10 over LSOIE, but $P$ goes down by exactly that much - though F1 does improve. To test the TANL model, we try only one linguistic feature, PoS. Like the ClausIE-extracted dataset, this dataset improves much better on recall, but unlike ClausIE, also shows a modest improvement on $P$. TANL highly relies on a fully correct sequence generation. However, OIE multi-tuple extraction for long sentences can get very long to correctly generate in one shot. Thus, we break it down by tagging each verb (for a possible predicate proxy) to generate all extractions, with multi-tuple facts possible for each predicate. This forms the primary reason for the low $P$ - because all predicates are tagged, we get many extractions for a sentence, many of which may be spurious. Besides, due to the tags appearing as full words instead of symbols, this further distances facts by elongating the tuple, which makes room for more mistakes. However, it's exciting to note that the same strategy contributes to the large jump in recall - because now each predicate is being focused on every input, a larger extent of the expected facts can be covered.

Finally, fine-tuning models trained for multiple SP tasks (like TANL) seem to help in OIE. However, OIE still seems harder than other SP tasks - the tagging strategy seems to be the primary influence over scores, rather than the pre-training itself. Future work can better help quantify TANL (or DeepStruct) on OIE performance.

\begin{table}[h]
\centering
\resizebox{\columnwidth}{!}{%
\begin{tabular}{c|ccc}
\textbf{TANL Format}   & \textbf{P} & \textbf{R} & \textbf{F1} \\ \hline
Plain $wt_{src}=1$,        & 40.5                  & 36.1                  & 38.2                   \\ \hdashline
PoS $wt_{src}=0.6$, $wt_{PoS}=0.4$   & 42.4                  & 39.6                  & 40.9                   \\
PoS $dim_{src}=768$, $dim_{PoS}=30$ & \textbf{44.7}         & \textbf{43.3}         & \textbf{44.0}         
\end{tabular}
}
\label{TANL_SCORES_LSOIE}
\caption{P, R and F1 scores for fine-tuning on TANL. Trained on the LSOIE-extracted dataset using the TANL-format.}
\end{table}

\textbf{How do WA and LC perform with respect to each other?} We note that both have advantages in differing settings. For example, WA performs beter for the LSOIE-extracted dataset, but LC performs better for its TANL-format. For the ClausIE-extracted dataset, there is little difference in performance boost.
Hence, although both methods seem mathematically similar, we experimentally validate the need for both types.

% WAvg: (cie:  0.4/-0.3 +9, +6). (lsoie: +12ish, +3, +5)
%         tanl: 2,3.5, 2.9
% LC: (cie: -2, 9.4, +6.1), (lsoie: +9, 1.6, 3.8)
%         tanl: 4, 7.2, 5.8

\textbf{Which linguistic features better help the model?} There seems to be no clear winner among the three features. Each perform better than the other in different settings, showing the utility of all tags. Interestingly, combinations of PoS and SynDP didn't seem to improve the model. Hence, we also do not experiment with including the head word/tag in SynDP.

\textbf{How does the new SemDP feature fare?} We bring to attention that SemDP shows as much improvement as other features, cementing it's usefulness. Infact, this improvement came with just using the first tag of each word, meaning most such tags were just blank '\_' tokens. Multiple tags (of the other heads that that word depends on) may strengthen performance, which can be investigated in the future. Because SemDP uses the smallest tagset - with just about three tags being most frequently used - to gain equivalent performance, it could mean lesser training time and lower energy cost. Thus, we consider SemDP to be the best amongst the features.

% PoS vs SemDP on WA Cie: slightly favours PoS. Infact combo is lowest.

% PoS vs SynDP on WA LSOIE: slight favour towards SynDP. Combo shows highest $P$ among the three but lowest R, but stil SynDP doing better by 0.1.

% PoS vs SemDP on LC Cie: SLight favour of SemDP around 30, but for 10 PoS more. Combo is strongest here.
% Last tabl favours PoS over Sydp

% 

\section{Conclusion}
\label{conclusion}
We propose using two novel techniques for enhancing word embeddings with their linguistic features (PoS, SynDP and SemDP) - demonstrated for the OIE task with a generative approach. Both techniques, Weighted Addition and Linearized Concatenation exhibit vast improvements over the dataset baselines, and each shows importance in different settings. We also show how SemDP can be used successfully as a linguistic feature, which has not been explored before. We also extend the TANL format for OIE, and contribute a better quality synthetic dataset made from ClausIE. We also contribute the first benchmarks for models trained generatively on a Seq2Seq PLM with the LSOIE dataset.

\section{Limitations and Future Directions}
\label{limitations}
The usefulness of including linguistic features highly depends on how accurate such taggers are.  We point out that a better study of available linguistic annotating tools and how they affect performance needs to be carried out in the context of OIE. As noted earlier, OPIEC makes for a valuable resource but its extractions come from MinIE, whose tuple-annotated format largely differs from current triple formats.

The predicate-tagging strategy developed hurts precision but increases recall. A better way that either limits the predicate tags or proposes a novel strategy for generative OIE needs to be explored to help multi-tuple extractions.

Our SemDP and SynDP information only considers the linguistic tag. To further test previous works' conclusions, it would be useful to experiment with combining the embeddings of the head word or tag too.

As our proposed embedding approaches can be used regardless of whether the OIE system is tagging-based or generative, it would be important to explore various tagging-based models with our proposed improvements. 

We understand that it is necessary to check if other OIE models benefit from our methods. However, this would mean configuring each model with their host of dependency requirements, checking how each system produces output and again directing that to the format benchmarks expect. Due to space constraints, we believe that it would be better to keep this investigation as extended work as we work on almost exhaustively covering all aspects of our model. A future more in-length comparison can then allow for detailed and patient investigations that would be overwhelming to present here.

As OIE is a structured prediction task, it would be important to explore our embedding approach for other such tasks, and even more so if the model is being trained on all structured prediction tasks, like TANL. As this approach is not dependent on the task itself, future work can benchmark structural utility for other NLP tasks, like translation and question answering.

Noting our low recall scores by subjectively analysing LSOIE, we highlight an important need for higher-quality training datasets for OIE, a limitation for any model's performance. We call for standardising datasets to triples instead of $n$-ary tuples, which would make OIE more useful for downstream tasks (such as knowledge base creation). Besides, we notice that OIE systems continue to be benchmarked on differing evaluation scorers. Publishing precision, recall and F1 scores in atleast one common golden framework helps compare various models quickly, which is vital to research.

DeepStruct is an alternative model to TANL, but doesn't benchmark on better standards like CaRB. Due to their final code not giving instructions for OIE, we weren't able to explore this model. Its usefulness for OIE needs to be evaluated to better contribute to the study of the SP-OIE relationship. 
\nocite{*}

\section{Acknowledgements}
We thank our colleagues at IvLabs (the AI \& Robotics Lab of our institution) for their support and ancillary compute resources. The first three authors are associated with this lab and acknowledge the instrumental role it has played for us. \\
We thank the reviewers for their time and feedback.

\section{References}
\label{sec:reference}

\bibliographystyle{lrec-coling2024-natbib}
\bibliography{custom}

\begin{thebibliography}{32}
\expandafter\ifx\csname natexlab\endcsname\relax\def\natexlab#1{#1}\fi

\bibitem[{Angeli et~al.(2015{\natexlab{a}})Angeli, Johnson~Premkumar, and Manning}]{angeli-etal-2015-leveraging}
Gabor Angeli, Melvin~Jose Johnson~Premkumar, and Christopher~D. Manning. 2015{\natexlab{a}}.
\newblock \href {https://doi.org/10.3115/v1/P15-1034} {Leveraging linguistic structure for open domain information extraction}.
\newblock In \emph{Proceedings of the 53rd Annual Meeting of the Association for Computational Linguistics and the 7th International Joint Conference on Natural Language Processing (Volume 1: Long Papers)}, pages 344--354, Beijing, China. Association for Computational Linguistics.

\bibitem[{Angeli et~al.(2015{\natexlab{b}})Angeli, Premkumar, and Manning}]{angeli2015leveraging}
Gabor Angeli, Melvin Jose~Johnson Premkumar, and Christopher~D Manning. 2015{\natexlab{b}}.
\newblock Leveraging linguistic structure for open domain information extraction.
\newblock In \emph{Proceedings of the 53rd Annual Meeting of the Association for Computational Linguistics and the 7th International Joint Conference on Natural Language Processing (Volume 1: Long Papers)}, pages 344--354.

\bibitem[{Bhardwaj et~al.(2019)Bhardwaj, Aggarwal, and Mausam}]{bhardwaj-etal-2019-carb}
Sangnie Bhardwaj, Samarth Aggarwal, and Mausam Mausam. 2019.
\newblock \href {https://doi.org/10.18653/v1/D19-1651} {{C}a{RB}: A crowdsourced benchmark for open {IE}}.
\newblock In \emph{Proceedings of the 2019 Conference on Empirical Methods in Natural Language Processing and the 9th International Joint Conference on Natural Language Processing (EMNLP-IJCNLP)}, pages 6262--6267, Hong Kong, China. Association for Computational Linguistics.

\bibitem[{Bird(2006)}]{bird-2006-nltk}
Steven Bird. 2006.
\newblock \href {https://doi.org/10.3115/1225403.1225421} {{NLTK}: The {N}atural {L}anguage {T}oolkit}.
\newblock In \emph{Proceedings of the {COLING}/{ACL} 2006 Interactive Presentation Sessions}, pages 69--72, Sydney, Australia. Association for Computational Linguistics.

\bibitem[{Christensen et~al.(2011)Christensen, Mausam, Soderland, and Etzioni}]{10.1145/1999676.1999697}
Janara Christensen, Mausam, Stephen Soderland, and Oren Etzioni. 2011.
\newblock \href {https://doi.org/10.1145/1999676.1999697} {An analysis of open information extraction based on semantic role labeling}.
\newblock In \emph{Proceedings of the Sixth International Conference on Knowledge Capture}, K-CAP '11, page 113–120, New York, NY, USA. Association for Computing Machinery.

\bibitem[{Cui et~al.(2018)Cui, Wei, and Zhou}]{cui-etal-2018-neural}
Lei Cui, Furu Wei, and Ming Zhou. 2018.
\newblock \href {https://doi.org/10.18653/v1/P18-2065} {Neural open information extraction}.
\newblock In \emph{Proceedings of the 56th Annual Meeting of the Association for Computational Linguistics (Volume 2: Short Papers)}, pages 407--413, Melbourne, Australia. Association for Computational Linguistics.

\bibitem[{Del~Corro and Gemulla(2013)}]{del2013clausie}
Luciano Del~Corro and Rainer Gemulla. 2013.
\newblock Clausie: clause-based open information extraction.
\newblock In \emph{Proceedings of the 22nd international conference on World Wide Web}, pages 355--366.

\bibitem[{Dozat and Manning(2018)}]{dozat-manning-2018-simpler}
Timothy Dozat and Christopher~D. Manning. 2018.
\newblock \href {https://doi.org/10.18653/v1/P18-2077} {Simpler but more accurate semantic dependency parsing}.
\newblock In \emph{Proceedings of the 56th Annual Meeting of the Association for Computational Linguistics (Volume 2: Short Papers)}, pages 484--490, Melbourne, Australia. Association for Computational Linguistics.

\bibitem[{Du et~al.(2022)Du, Qian, Liu, Ding, Qiu, Yang, and Tang}]{du-etal-2022-glm}
Zhengxiao Du, Yujie Qian, Xiao Liu, Ming Ding, Jiezhong Qiu, Zhilin Yang, and Jie Tang. 2022.
\newblock \href {https://doi.org/10.18653/v1/2022.acl-long.26} {{GLM}: General language model pretraining with autoregressive blank infilling}.
\newblock In \emph{Proceedings of the 60th Annual Meeting of the Association for Computational Linguistics (Volume 1: Long Papers)}, pages 320--335, Dublin, Ireland. Association for Computational Linguistics.

\bibitem[{Gashteovski et~al.(2017)Gashteovski, Gemulla, and del Corro}]{gashteovski-etal-2017-minie}
Kiril Gashteovski, Rainer Gemulla, and Luciano del Corro. 2017.
\newblock \href {https://doi.org/10.18653/v1/D17-1278} {{M}in{IE}: Minimizing facts in open information extraction}.
\newblock In \emph{Proceedings of the 2017 Conference on Empirical Methods in Natural Language Processing}, pages 2630--2640, Copenhagen, Denmark. Association for Computational Linguistics.

\bibitem[{Gashteovski et~al.(2019)Gashteovski, Wanner, Hertling, Broscheit, and Gemulla}]{gashteovski2019opiec}
Kiril Gashteovski, Sebastian Wanner, Sven Hertling, Samuel Broscheit, and Rainer Gemulla. 2019.
\newblock Opiec: An open information extraction corpus.
\newblock In \emph{Proceedings of the Conference on Automatic Knowledge Base Construction (AKBC)}.

\bibitem[{Kingma and Ba(2014)}]{kingma2014adam}
Diederik~P Kingma and Jimmy Ba. 2014.
\newblock Adam: A method for stochastic optimization.
\newblock \emph{arXiv preprint arXiv:1412.6980}.

\bibitem[{Kolluru et~al.(2020)Kolluru, Aggarwal, Rathore, {Mausam}, and Chakrabarti}]{kolluru-etal-2020-imojie}
Keshav Kolluru, Samarth Aggarwal, Vipul Rathore, {Mausam}, and Soumen Chakrabarti. 2020.
\newblock \href {https://doi.org/10.18653/v1/2020.acl-main.521} {{IM}o{JIE}: Iterative memory-based joint open information extraction}.
\newblock In \emph{Proceedings of the 58th Annual Meeting of the Association for Computational Linguistics}, pages 5871--5886, Online. Association for Computational Linguistics.

\bibitem[{Lacoste et~al.(2019)Lacoste, Luccioni, Schmidt, and Dandres}]{lacoste2019quantifying}
Alexandre Lacoste, Alexandra Luccioni, Victor Schmidt, and Thomas Dandres. 2019.
\newblock Quantifying the carbon emissions of machine learning.
\newblock \emph{arXiv preprint arXiv:1910.09700}.

\bibitem[{Lewis et~al.(2020)Lewis, Liu, Goyal, Ghazvininejad, Mohamed, Levy, Stoyanov, and Zettlemoyer}]{lewis-etal-2020-bart}
Mike Lewis, Yinhan Liu, Naman Goyal, Marjan Ghazvininejad, Abdelrahman Mohamed, Omer Levy, Veselin Stoyanov, and Luke Zettlemoyer. 2020.
\newblock \href {https://doi.org/10.18653/v1/2020.acl-main.703} {{BART}: Denoising sequence-to-sequence pre-training for natural language generation, translation, and comprehension}.
\newblock In \emph{Proceedings of the 58th Annual Meeting of the Association for Computational Linguistics}, pages 7871--7880, Online. Association for Computational Linguistics.

\bibitem[{{Mausam} et~al.(2012){Mausam}, Schmitz, Soderland, Bart, and Etzioni}]{mausam-etal-2012-open}
{Mausam}, Michael Schmitz, Stephen Soderland, Robert Bart, and Oren Etzioni. 2012.
\newblock \href {https://aclanthology.org/D12-1048} {Open language learning for information extraction}.
\newblock In \emph{Proceedings of the 2012 Joint Conference on Empirical Methods in Natural Language Processing and Computational Natural Language Learning}, pages 523--534, Jeju Island, Korea. Association for Computational Linguistics.

\bibitem[{Miller et~al.(1998)Miller, Crystal, Fox, Ramshaw, Schwartz, Stone, and Weischedel}]{10.3115/1119089.1119107}
Scott Miller, Michael Crystal, Heidi Fox, Lance Ramshaw, Richard Schwartz, Rebecca Stone, and Ralph Weischedel. 1998.
\newblock \href {https://doi.org/10.3115/1119089.1119107} {Algorithms that learn to extract information: Bbn: Tipster phase iii}.
\newblock In \emph{Proceedings of a Workshop on Held at Baltimore, Maryland: October 13-15, 1998}, TIPSTER '98, page 75–89, USA. Association for Computational Linguistics.

\bibitem[{Mtumbuka and Lukasiewicz(2022)}]{mtumbuka-lukasiewicz-2022-syntactically}
Frank Mtumbuka and Thomas Lukasiewicz. 2022.
\newblock \href {https://aclanthology.org/2022.emnlp-main.401} {Syntactically rich discriminative training: An effective method for open information extraction}.
\newblock In \emph{Proceedings of the 2022 Conference on Empirical Methods in Natural Language Processing}, pages 5972--5987, Abu Dhabi, United Arab Emirates. Association for Computational Linguistics.

\bibitem[{Niklaus et~al.(2018)Niklaus, Cetto, Freitas, and Handschuh}]{niklaus2018survey}
Christina Niklaus, Matthias Cetto, Andr{\'e} Freitas, and Siegfried Handschuh. 2018.
\newblock A survey on open information extraction.
\newblock \emph{arXiv preprint arXiv:1806.05599}.

\bibitem[{Oepen et~al.(2019)Oepen, Abend, Hajic, Hershcovich, Kuhlmann, O{'}Gorman, Xue, Chun, Straka, and Uresova}]{oepen-etal-2019-mrp}
Stephan Oepen, Omri Abend, Jan Hajic, Daniel Hershcovich, Marco Kuhlmann, Tim O{'}Gorman, Nianwen Xue, Jayeol Chun, Milan Straka, and Zdenka Uresova. 2019.
\newblock \href {https://doi.org/10.18653/v1/K19-2001} {{MRP} 2019: Cross-framework meaning representation parsing}.
\newblock In \emph{Proceedings of the Shared Task on Cross-Framework Meaning Representation Parsing at the 2019 Conference on Natural Language Learning}, pages 1--27, Hong Kong. Association for Computational Linguistics.

\bibitem[{Paolini et~al.(2021)Paolini, Athiwaratkun, Krone, Ma, Achille, Anubhai, dos Santos, Xiang, and Soatto}]{tanl}
Giovanni Paolini, Ben Athiwaratkun, Jason Krone, Jie Ma, Alessandro Achille, Rishita Anubhai, Cicero~Nogueira dos Santos, Bing Xiang, and Stefano Soatto. 2021.
\newblock Structured prediction as translation between augmented natural languages.
\newblock In \emph{9th International Conference on Learning Representations, {ICLR} 2021}.

\bibitem[{Qi et~al.(2020)Qi, Zhang, Zhang, Bolton, and Manning}]{qi2020stanza}
Peng Qi, Yuhao Zhang, Yuhui Zhang, Jason Bolton, and Christopher~D Manning. 2020.
\newblock Stanza: A python natural language processing toolkit for many human languages.
\newblock \emph{arXiv preprint arXiv:2003.07082}.

\bibitem[{Raffel et~al.(2020)Raffel, Shazeer, Roberts, Lee, Narang, Matena, Zhou, Li, and Liu}]{JMLR:v21:20-074}
Colin Raffel, Noam Shazeer, Adam Roberts, Katherine Lee, Sharan Narang, Michael Matena, Yanqi Zhou, Wei Li, and Peter~J. Liu. 2020.
\newblock \href {http://jmlr.org/papers/v21/20-074.html} {Exploring the limits of transfer learning with a unified text-to-text transformer}.
\newblock \emph{Journal of Machine Learning Research}, 21(140):1--67.

\bibitem[{Roy et~al.(2019)Roy, Park, Lee, and Pan}]{roy-etal-2019-supervising}
Arpita Roy, Youngja Park, Taesung Lee, and Shimei Pan. 2019.
\newblock \href {https://doi.org/10.18653/v1/D19-1067} {Supervising unsupervised open information extraction models}.
\newblock In \emph{Proceedings of the 2019 Conference on Empirical Methods in Natural Language Processing and the 9th International Joint Conference on Natural Language Processing (EMNLP-IJCNLP)}, pages 728--737, Hong Kong, China. Association for Computational Linguistics.

\bibitem[{Saha and {Mausam}(2018)}]{saha-mausam-2018-open}
Swarnadeep Saha and {Mausam}. 2018.
\newblock \href {https://aclanthology.org/C18-1194} {Open information extraction from conjunctive sentences}.
\newblock In \emph{Proceedings of the 27th International Conference on Computational Linguistics}, pages 2288--2299, Santa Fe, New Mexico, USA. Association for Computational Linguistics.

\bibitem[{Solawetz and Larson(2021)}]{solawetz-larson-2021-lsOIE}
Jacob Solawetz and Stefan Larson. 2021.
\newblock \href {https://doi.org/10.18653/v1/2021.eacl-main.222} {{LSOIE}: A large-scale dataset for supervised open information extraction}.
\newblock In \emph{Proceedings of the 16th Conference of the European Chapter of the Association for Computational Linguistics: Main Volume}, pages 2595--2600, Online. Association for Computational Linguistics.

\bibitem[{Stanovsky and Dagan(2016)}]{stanovsky2016creating}
Gabriel Stanovsky and Ido Dagan. 2016.
\newblock Creating a large benchmark for open information extraction.
\newblock In \emph{Proceedings of the 2016 Conference on Empirical Methods in Natural Language Processing}, pages 2300--2305.

\bibitem[{Stanovsky et~al.(2018)Stanovsky, Michael, Zettlemoyer, and Dagan}]{stanovsky-etal-2018-supervised}
Gabriel Stanovsky, Julian Michael, Luke Zettlemoyer, and Ido Dagan. 2018.
\newblock \href {https://doi.org/10.18653/v1/N18-1081} {Supervised open information extraction}.
\newblock In \emph{Proceedings of the 2018 Conference of the North {A}merican Chapter of the Association for Computational Linguistics: Human Language Technologies, Volume 1 (Long Papers)}, pages 885--895, New Orleans, Louisiana. Association for Computational Linguistics.

\bibitem[{Wang et~al.(2022)Wang, Liu, Chen, Hong, Tang, and Song}]{wang-etal-2022-deepstruct}
Chenguang Wang, Xiao Liu, Zui Chen, Haoyun Hong, Jie Tang, and Dawn Song. 2022.
\newblock \href {https://doi.org/10.18653/v1/2022.findings-acl.67} {{D}eep{S}truct: Pretraining of language models for structure prediction}.
\newblock In \emph{Findings of the Association for Computational Linguistics: ACL 2022}, pages 803--823, Dublin, Ireland. Association for Computational Linguistics.

\bibitem[{Yates et~al.(2007)Yates, Cafarella, Banko, Etzioni, Broadhead, and Soderland}]{10.5555/1614164.1614177}
Alexander Yates, Michael Cafarella, Michele Banko, Oren Etzioni, Matthew Broadhead, and Stephen Soderland. 2007.
\newblock Textrunner: Open information extraction on the web.
\newblock In \emph{Proceedings of Human Language Technologies: The Annual Conference of the North American Chapter of the Association for Computational Linguistics: Demonstrations}, NAACL-Demonstrations '07, page 25–26, USA. Association for Computational Linguistics.

\bibitem[{Zhan and Zhao(2020)}]{zhan2020span}
Junlang Zhan and Hai Zhao. 2020.
\newblock Span model for open information extraction on accurate corpus.
\newblock In \emph{Proceedings of the AAAI Conference on Artificial Intelligence}, volume~34, pages 9523--9530.

\bibitem[{Zhou et~al.(2022)Zhou, Yu, Sun, Long, Li, and Sun}]{zhou2022survey}
Shaowen Zhou, Bowen Yu, Aixin Sun, Cheng Long, Jingyang Li, and Jian Sun. 2022.
\newblock A survey on neural open information extraction: Current status and future directions.
\newblock \emph{arXiv preprint arXiv:2205.11725}.

\end{thebibliography}

\appendix 

\section{Ablations}
\label{Appendix_Ablations}
\subsection{Dataset size and domain}
We experiment our PoS enhancement approach by scaling it to a much larger dataset, that combines both the Wiki and Science domains. Though more data is preferred for neural networks, we observe that training on such a specific domain actually hurts the plain model (Table \ref{full-vs-wiki}) However, with both our embedding frameworks, we are able to turn this around and actually help the model perform atleast on par with the wiki-only dataset. Further performance is possibly hindered due to the dataset quality and imbalance in the domain ratio of the original LSOIE dataset: Wiki has around 24k total sentences and Sci has around 48k total. In our Seq2Seq processed versions of the LSOIE dataset, we had around 22k and 45k total sentences respectively for Wiki and Sci.

\begin{table}[h]
\centering
\resizebox{\columnwidth}{!}{%
\begin{tabular}{l|llll}
\textbf{Dataset}                                 & \textbf{Model}                & \textbf{P}             & \textbf{R}             & \textbf{F1}            \\ \hline
\multirow{3}{*}{LSOIE-(wiki)}        & Plain $wt_{src}=1$        & 50.5          & 26.1          & 34.4          \\
                                        & $wt_{src}=0.6, wt_{PoS}=0.4$  & \textbf{60.2} & \textbf{29}   & \textbf{39.1} \\
                                        & $dim_{src}=768, dim_{PoS=30}$ & 59.5          & 27.7          & 37.8          \\ \hline
\multirow{3}{*}{LSOIE-(wiki+sci)} & Plain $wt_{src}=1$        & 48.7          & 26.0          & 33.9          \\
                                        & $wt_{src}=0.6, wt_{PoS}=0.4$  & \textbf{60.6} & 28.8          & \textbf{39.1} \\
                                        & $dim_{src}=768, dim_{PoS=30}$ & 59.7          & \textbf{28.9} & 38.9         
\end{tabular}
}
\caption{CaRB scores comparing wiki+sci and wiki splits of our Seq2Seq processed version of LSOIE.}
\label{full-vs-wiki}
\end{table}

\subsection{Frozen source embeddings}
We have allowed our model to update its pre-trained source embeddings, and now compare freezing the input source embeddings from the T5 model. This allows only the feature tags to be trained with respect to the original source, while preserving the pre-trained meaning in the source embedding. As can be seen in Table \ref{freeze}, freezing doesn't help the model, except for marginally improved precision. This demonstrates the importance of actually enhancing the source word embedding with respect to its structural information.

\begin{table}[h]
\resizebox{\columnwidth}{!}{%
\begin{tabular}{l|lll}
\textbf{Model}                    & \textbf{P}    & \textbf{R}    & \textbf{F1}   \\ \hline
$wt_{src}=0.6, wt_{PoS}=0.4$      & 60.2 & \textbf{29}   & \textbf{39.1} \\
$wt_{src}=0.6, wt_{PoS}=0.4$ \textit{(F)}  & \textbf{60.6} & 28.1 & 38.4 \\ \hline
$dim_{src}=768, dim_{PoS}=30$     & 59.5 & \textbf{27.7} & \textbf{37.8} \\
$dim_{src}=768, dim_{PoS}=30$ \textit{(F)} & \textbf{59.9} & 27.3 & 37.5
\end{tabular}
}
\caption{CaRB scores (trained on the LSOIE-extracted dataset) for freezing pre-trained word embeddings (denoted by $F$) compared to allowing the model to fine-tune them along with PoS.} 
\label{freeze}
\end{table}

\section{Task-specific prefix}
For fine-tuning the pre-trained T5 model, we follow the “text-to-text framework” used by the authors of T5 to fine-tune their T5 models. For each task (translation, question answering etc.), they add a “task-specific prefix” (text)  to the original input text before feeding it to the T5 model. For example, to translate from English to French, the model would be given the sequence “\texttt{translate English to French: }” followed by the actual sentence. As our task focuses on OIE, we use the prefix “\texttt{info\_extract: }” for each input sequence. Naturally, we do not want the model to associate linguistic features with the text in the task-specific prefix. Keeping this in mind, we kept the \texttt{<pad>} tag for the tokens in the prefix and used the linguistic features of the actual sentence. 

\section{Experimental Details}
\label{Experimental_Details}
We train all the models with the Adam \cite{kingma2014adam} optimizer with a learning rate of 1e-4. We train all the models for 15 epochs and the best models are chosen based on the validation set results.
The hyperparameters were determined over a set of preliminary experiments, and were kept constant throughout the experiments. These are summarized in Table \ref{hyperparams}. We use a batch size of 32, and the training time per epoch was around 5 - 7 minutes. 

\subsection{Carbon Emissions from Experiments} Using Electricity Maps\footnote{\url{https://app.electricitymaps.com}} , an estimate for the carbon efficiency was obtained. Our zone is Western India, of which we chose to use the average of the last 12 month's consumption factor, 0.713 kgCO$_2$eq/kWh.

Around 50 hours of experiments were conducted on a single Titan RTX (TDP of 280W). Total emissions are estimated to be 2.99 kgCO$_2$eq.

Estimations were conducted using the MachineLearning Impact calculator\footnote{\url{https://mlco2.github.io/impact\#compute}} \cite{lacoste2019quantifying}.

\begin{table}[h]
\centering
\begin{tabular}{|l|l|l|}
\hline
\textbf{Hyperparameter}   & \textbf{Value}\\\hline
Learning Rate & 1e-4\\
Batch Size & 32 \\
Epochs trained & 15\\
Max. Token Length & 64\\\hline
\end{tabular}
\caption{Hyperparameters}
\label{hyperparams}
\end{table}

\begin{table*}[h]
\centering
\begin{tabular}{|p{0.1325\linewidth}|p{0.1825\linewidth}|p{0.3325\linewidth}|p{0.2825\linewidth}|}
\hline
\textbf{Token} &
  \textbf{Part-of-Speech (PoS)} &
  \textbf{Syntactic Dependency Parse (SynDP)} &
  \textbf{Semantic Dependency Parse (SemDP)}
   \\ \hline
info\_extract  & <pad> & <pad> & <pad> \\ \hline
: & <pad> & <pad> & <pad> \\ \hline
The &
 DT (determiner) &
   study: det (determiner) & \_
   \\ \hline
study &
  NN (common noun) &
  published: nsubj:pass (passive nominal subject) & The: BV | published: ARG2 | Change: ARG1
   \\ \hline
was &
  VBD (past tense verb) &
   published: aux:pass (passive auxiliary) & \_
   \\ \hline
   published &
  VBN (past participle verb) &
   0: root & 0: root | in: ARG1 | yesterday: loc
   \\ \hline
   in &
  IN (preposition) &
   journal: case (case marking) & \_
   \\ \hline
   journal &
  JJ (adjective) &
   published: obl (oblique nominal) & \_
   \\ \hline
     Nature &
  NNP (singular proper noun) &
   Change: compound & \_
   \\ \hline
     Climate &
  NNP &
   Change: compound & \_
   \\ \hline
     Change &
  NNP &
   journal: appos (appositional modifier) & 0: root | in: ARG2 | Climate: compound | yesterday: loc
   \\ \hline
     yesterday &
 NN &
   published: obl:tmod (temporal modifier) & \_
   \\ \hline
   . &
  . (sentence terminator) &
   published: punct (punctuation) & \_
   \\ \hline
   </s> (EOS token) & </s> & </s> & </s> \\ \hline
   
\end{tabular}
\caption{Example of linguistic tags for the sentence \textit{"The study was published in journal Nature Climate Change yesterday."} (Token indices were replaced by the actual token in the SynDP and SemDP tags)}
\label{linguistic-ftrs-ex}
\end{table*}

\section{Linguistic features}
\label{Linguistic_Features}
\subsection{Part of Speech}
Part of Speech (PoS) categorizes words that have similar syntactic and grammatical functions in a sentence. Familiar examples of these word classes in English include nouns, verbs and adjectives. We use the Stanza PoS tagger, which uses the UPenn TreeBank PoS Tagset. The complete list of tags and their descriptions can easily be seen by running the code using NLTK \cite{bird-2006-nltk}:

\begin{lstlisting}
    import nltk
    nltk.download('tagsets')
    nltk.help.upenn_tagset()
\end{lstlisting}

More about PoS tagging can be found here:
\begin{itemize}
    \item \url{https://en.wikipedia.org/wiki/Part_of_speech}
    \item \url{https://en.wikipedia.org/wiki/Part-of-speech_tagging}
    \item \url{https://www.nltk.org/book/ch05.html}
\end{itemize}

\begin{table*}[h]
\centering
\begin{tabular}{|p{0.31\linewidth}|p{0.31\linewidth}|p{0.31\linewidth}|}
\hline
\textbf{LSOIE Sentence} &
  \textbf{LSOIE Labels} &
  \textbf{Generated Labels}
   \\ \hline
Akerson will also relinquish his chairman role, to be replaced by current director Theodore Solso. &
  (Akerson;will relinquish;his chairman role) &
  (Akerson;will relinquish;his chairman role) (current director Theodore Solso;will replaced;Akerson)
   \\ \hline
Road accidents killed 8,600 on the nation's roads last year. &
  (on the nation's roads last year;killed;8,600) &
  (Road accidents;killed;8,600 on the nation's roads last year)
   \\ \hline
He said the world and the Paralympic movement is aware of the situation in the Ukraine, but the IPC needs to stay true to its mission. &
   (the IPC;needs;to stay true to its mission) (the IPC;stay;true to its mission) &
   (the IPC;should stay;true to its mission)
   \\ \hline
\end{tabular}
\caption{Examples where the original LSOIE (test) data is not clean and how our model (trained on the LSOIE-extracted dataset) gives better extractions despite being trained on it}
\label{tab-lsOIEissues}
\end{table*}

\subsection{Syntactic Dependency Parsing}
SynDP tags describe the syntactic relationships amongst words in a sentence, such as which words are subjects or objects of which clause. A token can be a "dependent" of only one "head" token. The relations are thus seen as directed edges - for example, the PoS for "The" in the table is "Determiner", while the SynDP tag specifies that it is the determiner for the word "study". A root token is identified that helps visualize the words as a rooted tree. We use Stanza's SynDP tagger, which uses the framework set by Universal Dependencies (UD). The complete tagset and their descriptions can be found under the alphabetical listing here: \url{https://universaldependencies.org/u/dep/index.html}

Details about SynDP can be found here:
\begin{itemize}
    \item \url{https://web.stanford.edu/~jurafsky/slp3/old_oct19/15.pdf}
    \item \url{https://courses.grainger.illinois.edu/cs447/fa2019/Slides/Lecture19.pdf}
    \item \url{https://stanfordnlp.github.io/stanza/depparse.html#accessing-syntactic-dependency-information}
\end{itemize}

\subsection{Semantic Dependency Parsing}
SemDP focuses on identifying meaningful relations between words. Unlike SynDP, not all words are tagged, because many words only contribute to the grammatical structure but aren't important to the meaning of the sentence. Infact, most tokens end up not being tagged, but the ones that are tagged are allowed to have multiple heads (unlike SynDP), including those that don't have a tag.
SemDP has a much smaller tagset need than SynDP. Most tags are usually ARG1 and ARG2 (similar to subj/obj), meant to show the importance of the connection of those words directly - without verbose syntactic formality. For example, in the table, "study" is semantically connected to both "published" (as ARG2) and to the journal's name (here, the final compounded token "Change" as ARG1). However, in SynDP, "study" and "journal" can only meet at the common head "published", which indirectly shows the link between the two,

We use SuPar for SemDP, which in turn uses the DM tagset. We believe that SemDP hasn't received as much attention as SynDP has. Infact, although there is research literature on improving SemDP parsers, there are far fewer resources attempting to explain its tagsets to a layman. However, we endeavour to atleast make a list of the DM tagset available to the reader, but couldn't find official explanations for them. However, their use can be easily understood in context when parsing various example sentences. The DM tagset is as follows: \texttt{ARG1, ARG2, compound, BV, root, poss, loc, -and-c, ARG3, times, mwe, appos, conj, neg, subord, -or-c, -but-c, \_}

SuPar follows the parser developed by \cite{dozat-manning-2018-simpler}. A good description of the three common SDP tagging frameworks can be found in Section 3 of the Proceedings of the Shared Task on Cross-Framework MRP at the 2019 CONLL \cite{oepen-etal-2019-mrp}

\subsection{Additional tags for T5}
In addition to the linguistic tags mentioned earlier, we also three tags that go hand in hand with the special subword tokens of T5. They are:

\begin{itemize}
    \item \textbf{\texttt{<pad>}}: Used when the token at hand is a \texttt{<pad>} token or is the task-specific prefix as mentioned earlier.
    \item \textbf{\texttt{<unk>}}: Used when the token at hand is a \texttt{<unk>} token.
    \item \textbf{\texttt{</s>}}: Used when the token at hand is an end-of sentence (\texttt{</s>}) token.
\end{itemize}

Unlike other models, T5 does not have a separate token to mark the beginning of sentence. The decoder is made to be aware of the semantics of the input sentence with the help of the task prefix itself.

\section{CaRB Evaluation}
\label{CaRB_Evaluation}
CaRB \citep{bhardwaj-etal-2019-carb} is an evaluation benchmark for OIE. The dataset was created by crowdsourcing manual annotations for OIE2016's \citep{stanovsky2016creating} sentences. It contributes a comprehensive evaluation framework that fairly takes into account the output styles of diverse OIE systems. CaRB matches relation with relation, and arguments with arguments, which is a much better evaluation criteria than some older benchmarks, such as OIE2016, which serializes the tuples into a sentence and just computes lexical matches. Though CaRB can give AUC and PR curves when confidence scores are provided for each extraction, we do not do so, as we use a generative architecture which doesn't give such direct tag scores.

We note that the LSOIE dataset allows $n$-ary tuples, but OIE benchmarks expect multiple triples to be extracted instead. This would cause models directly trained on LSOIE to score lower on CaRB than other models trained on triple-only datasets. 
% Due to this, we follow the pre-processing stage outlined in Section \ref{dataset}.

\section{Subjective Analysis of LSOIE}
\label{LSOIE_Issues}

Table \ref{tab-lsOIEissues} shows three sentences from the LSOIE test set. The first example shows that LSOIE's annotations don't teach the model to extract all candidate tuples, which heavily affects recall scores. However, our model has still learned to output both required tuples in this case. The second example shows that LSOIE has wrong extractions where the main subject itself was not extracted, affecting both precision and recall. Again, our model has still learnt to give a complete extraction. The third example shows the redundancy of multiple extractions in the dataset, and reinforces the issue of not extracting all candidate tuples. Such examples are common across the dataset, heavily affecting both precison and recall. In this case, our model has learnt to not be redundant, but has missed extracting information from the first part of the sentence, as such noisy training examples are frequent in LSOIE.

\section{ClausIE-extracted Dataset Creation}
\label{ClausIE_Creation}
For each source sentence in the LSOIE-extracted dataset, we obtain the OIE triples from the ClausIE extractor\footnote{\url{https://resources.mpi-inf.mpg.de/d5/clausie/}}. As a pre-processing step, we neglect the triples which are redundant. For example, for the sentence \textit{"She replaces Daniel Akerson , who was appointed by the government as both chief executive and chairman in 2009 during the company 's bankruptcy"} ClausIE's extractor gives 4 triples:
\begin{itemize}
    \item \textit{'(She; replaces; Daniel Akerson)'}
    \item \textit{'(the company; has; bankruptcy)'}
    \item \textit{'(Daniel Akerson; was appointed; by the government as both chief executive and chairman in 2009)'}
    \item \textit{'(Daniel Akerson; was appointed; by the government as both chief executive and chairman in 2009 during the company 's bankruptcy)'}
\end{itemize}
Clearly, the information present in the third triple is already contained in the fourth triple. For such cases, we only keep the triple which contains more information and remove the others. 

Additionally, a \texttt{NullPointerException} is raised by ClausIE's ClauseDetector for certain sentences, and its triples couldn't be extracted. We remove such sentences since their count was quite low compared to the entire dataset (~2\%). Table \ref{synthetic_dataset} summarizes the dataset statistics of the LSOIE and ClausIE extractions. Table \ref{tab:clausie-lsoie-comp} subjectively shows how the ClausIE-extracted triples are better than the LSOIE-extracted ones. 

\begin{table}[h]
\resizebox{\columnwidth}{!}{%
\begin{tabular}{l|cc}
Dataset             & \# train sent.       & \# valid sent.       \\ \hline
LSOIE-wiki (Original) & 19,625             & 2,402  \\
LSOIE-extracted (Seq2Seq)   & 18,100         & 2,086          \\
ClausIE-extracted (Seq2Seq) & 17,630         & 2,045          \\
\end{tabular}
}
\caption{Number of training and validation sentences in our Seq2Seq processed versions compared with the original LSOIE-wiki dataset that they were created from.}
\label{synthetic_dataset}
\end{table}

Furthermore, the PoS, SynDP and SemDP tags were extracted for each source sentence and added as part of the dataset. 
% Thus, our dataset has 6 columns - \texttt{[Source, LSOIE-extracted triples, ClausIE-extracted triples, SemDP tags, PoS tags, SynDP tags]}. 

\section{Extending TANL's format for OIE}
\label{Extending_TANL}

TANL was designed for various structure prediction tasks, but was not designed for OIE. We extended the TANL format for OIE. The input format was:
\textit{info\_extract: The cat [sat] on the mat.}

% Here, "info\_extract:" is the prefix to indicate that the task the model should perform is Open Information Extraction, as outlined earlier. 
The verb of each sentence in the input is tagged, to tell the model explicitly about the main verb of the sentence. If there are multiple verbs in the sentence, each is tagged separately, and the sentence is passed as input tagging a different verb each time to generate as many tuples as possible.

The output generated by the TANL is in the given format:
\textit{[[The cat | subject 1] [sat | predicate 1] [on the mat | object 1] | tuple 1]}

The output format is modelled after the other formats that TANL was originally trained on. The model will generate multiple tuples if possible, however, it is rare as the model is trained to generate the tuples associated with the tag verb, and usually one tag verb only results in one extraction.

\begin{table}[h]
\centering
\begin{tabular}{|p{0.465\linewidth}|p{0.465\linewidth}|}
\hline
\textbf{Input Sentence} &
  \textbf{LSOIE Label for TANL} \\ \hline
The new elections are {[}scheduled{]} to take place on February 2 of next year. &
  {[}{[}The new elections| subject 1{]} {[}scheduled| predicate 1{]} {[}to take place on February 2| object 1{]}| tuple1{]}{[}{[}The new elections| subject 2{]} {[}scheduled| predicate 2{]} {[}to take place February 2 of next year| object 2{]}| tuple2{]} \\ \hline
The new elections are scheduled to {[}take{]} place on February 2 of next year. &
  {[}{[}The new elections| subject 1{]} {[}will take| predicate 1{]} {[}place on February 2 of next year| object 1{]}| tuple1{]} \\ \hline
\end{tabular}%
\caption{TANL's input sentence format. The sentence is: "The new elections are scheduled to take place on February 2 of next year." Each verb is tagged separately with square brackets and the sentence is passed as input, tagging a different verb each time. The outputs are correspondingly generated and merged.}
\label{tab:tanl-input}
\end{table}

\begin{table*}[h]
\centering
\begin{tabular}{|p{0.31\linewidth}|p{0.31\linewidth}|p{0.31\linewidth}|}
\hline
LSOIE Sentence &
  LSOIE Labels &
  ClausIE Labels \\ \hline
Road accidents killed 8,600 on the nation 's roads last year . &
  (on the nation 's roads last year;killed;8,600) &
  (the nation; has; roads) (Road accidents; killed; 8,600 last year) (Road accidents; killed; 8,600 on the nation 's roads) \\ \hline
Approximately 45,000 service members are currently assigned to the military base , located in central Texas &
  (Approximately 45,000 service members;assigned;in central Texas) &
  (the military base; be located; in central Texas) (Approximately 45,000 service members; are assigned; to the military base located in central Texas currently) \\ \hline
\end{tabular}%
\caption{Examples where ClausIE-generated labels are better than LSOIE labels}
\label{tab:clausie-lsoie-comp}
\end{table*}

\end{document}